\documentclass{article}
\usepackage{spconf,amsmath,graphicx}
\usepackage[usenames,dvipsnames]{xcolor}

\usepackage{cite}
\usepackage{amsmath,amssymb,amsfonts,graphicx,bbold,amsthm,mathtools}
\usepackage{enumitem}
\usepackage{subfigure,epstopdf,algorithm}
\usepackage{array}
\usepackage{xcolor}
\usepackage{thmtools}

\usepackage{times}
\usepackage{eqparbox}
\usepackage[noend]{algpseudocode}

\makeatletter
\def\BState{\State\hskip-\ALG@thistlm}
\makeatother

	\title{Kernel Regression for Graph Signal Prediction in Presence of Sparse Noise}

	\name{Arun~Venkitaraman$^*$, Pascal Frossard$^\dagger$, and Saikat Chatterjee$^*$}
	\address{
		$^*$ Department of Information Science and Engineering,\\
		 School of Electrical Engineering and Computer Science,     
		KTH Royal Institute of Technology,  Sweden                 \\
		$^\dagger$ Signal Processing Laboratory (LTS4), Ecole Polytechnique F{\'e}d{\'e}rale de Lausanne (EPFL)\\
	arunv@kth.se,	pascal.frossard@epfl.ch, sach@kth.se
	}

\begin{document}
		\ninept
		\maketitle
	\begin{abstract}
	In presence of sparse noise we propose kernel regression for predicting output vectors which are smooth over a given graph. Sparse noise models the training outputs being corrupted either with missing samples or large perturbations. The presence of sparse noise is handled using appropriate use of $\ell_1$-norm along-with use of $\ell_2$-norm in a convex cost function. For optimization of the cost function, we propose an iteratively reweighted least-squares (IRLS) approach that is suitable for kernel substitution or kernel trick due to availability of a closed form solution. Simulations using real-world temperature data show efficacy of our proposed method, mainly for limited-size training datasets.
\end{abstract}
	

	\section{Introduction}
	 Kernel regression deals with the problem of learning a model relating a given input vector to its associated output vector, such that the learnt model provides a prediction for the output, when given an input. The model is learnt using a training dataset of input-output pairs and is based on a kernel function associating the different inputs. The kernel function captures the similarity between the various input points over a high-dimensional space. In the case when the  kernel function represents is given by the inner-product between input vectors, we get the more popular special case of linear regression\cite{Bishop}. Kernel regression plays a central role in a gamut of techniques from support vector machines \cite{Bishop}, extreme learning machines\cite{elm_HUANG2010} to deep learning\cite{kernel_deeplearning}. More recently, kernel-based approaches have been applied to the analysis of graph signals. 
	 
One of the earliest works which employ graph-regularization in conjunction with kernels is due to Smola et al. \cite{Smola2003}, though it was not in the context of graph signal processing. Kernel-based reconstruction of graph signals was proposed by Romero et al.  in \cite{kergraph1,kergraph2} where both the input and output are graph signal values at subsets of nodes of a given graph. This was then further extended to joint space-time graph signal reconstruction in \cite{kergraph3,kergraph4}. Along same lines of thought, Ionnidis et al. proposed a more general approach for inferring functions over graphs in both static and dynamic settings\cite{IOANNIDIS2018173}. Using the framework of diffusion wavelets, Chung et al employed kernel regression in mandible growth modeling in CT images\cite{CHUNG201563}. Shen et al. proposed the use of kernels in structural equation models for identifying network topologies from graph signals\cite{baingana1}. Kernel regression for graph signal outputs with inputs agnostic to a graph was considered first by Venkitaraman et al. \cite{Arun_kergraph}, which they later extended to its Bayesian version in the form of Gaussian processes over graphs\cite{Arun_GPG}. The related problem of using multi-kernel approaches which allow selection of kernels directly from data have also been pursued in the context of graph signals\cite{multikernel_2,Arun_mkrg}. \\
	 \indent However, most of the aforementioned works are formulated for either the case when there is no noise or that the noise is implicitly Gaussian distributed, which may not be always a realistic assumption. In many scenarios in practice, noise occurs usually at a sparse subset of nodes, either due to samples being missing or due to random large perturbations in the signal value. In such cases, the performance of these approaches becomes severely limited. Motivated by this observation, we propose kernel regression for predicting signals which are smooth over an associated graph, when the training data is scarce and noisy with the training outputs corrupted with sparse noise. By sparse noise, we mean the case where each training output may have an unknown subset of nodes with either missing samples or large perturbations in the signal value. In such adverse scenarios, applying kernel regression on each node individually would not yield a reasonable prediction performance. The use of structural information significantly boosts prediction performance in such case.  Since we consider outputs which are graph signals, we use the graph smoothness to improve prediction. \\
	 \indent
	We first start from the first principles of linear regression and learn the optimal regression coefficients by minimizing a cost consisting of both graph smoothness constraint and the $\ell_1$ norm of the model error, since we assume sparse noise. A direct minimization of the cost does not admit a closed-form solution necessary to arrive at the more general kernel based formulation. The kernel formulation is desirable over linear regression since it is usually unclear as to what is the best feature to be applied on the input used in linear regression. On the other hand, it is usually easier and more intuitive to express the relation (similarity/ dissimilarity) between various input observations using a kernel function. In order that the proposed linear regression further results in a kernel regression formulation, we take an iteratively reweighted least squares (IRLS) approach in solving the $\ell_1$-norm based optimization. 
%
	\section{Kernel Regression over Graphs in Presence of Sparse Noise}
	\label{sec:Kernel_Regression_over_Graphs}
	We first introduce the relevant background in graph signal processing and proceed to our problem formulation.
		\subsection{Brief review of graph signal processing basics}
	Consider a graph $G$ with $M$ nodes,  with the edge set $\mathcal{E}$ and the adjacency matrix $\mathbf{A}$, where $\textbf{A}(i,j)$ denotes the strength of the edge between the $i$th and $j$th nodes. We consider undirected graphs with symmetric edge-weights or $\mathbf{A}=\mathbf{A}^\top$. The graph-Laplacian matrix $\mathbf{L}$ of $G$ is given by
	$\mathbf{L=D-A}$, where $\mathbf{D}$ is the diagonal degree matrix with $i$th diagonal element given by the sum of the elements in the $i$th row of $\mathbf{A}$.  
	A graph signal $\mathbf{y}=[y(1)\,y(2)\,\cdots y(M)]^\top\in\mathbb{R}^M$ is a vector whose components are the values of a physical quantity at the various nodes of $\mathcal{G}$. The quadratic form 
	$\mathbf{y}^\top\mathbf{L}\mathbf{y}=\sum_{(i,j)\in\mathcal{E}} \mathbf{A}(i,j)(y(i)-y(j))^2$
	is usually used to quantify the smoothness of $\mathbf{y}$ over the graph: a small value indicates that $\mathbf{y}$ has similar values across connected nodes. Such a signal $\mathbf{y}$ is referred to as a smooth graph signal. Graph signal processing is a continuously expanding field of research, and we refer the reader to \cite{Shuman,gsp_overview_ortega} and the references therein for a more comprehensive study of the framework.

\subsection{Kernel Regression over Graphs in Presence of Sparse Noise}
	Let $\{\mathbf{x}_n\in\mathbb{R}^{N_i},\mathbf{t}_n\in\mathbb{R}^M\}_{n=1}^N$ denote the training dataset comprising $N$ input-output pairs. We consider outputs $\mathbf{t}_n$ which are smooth over a known graph of $M$ nodes with the graph-Laplacian matrix $\mathbf{L}$.  We assume that the training outputs are corrupted with sparse noise -- either through missing signal values or large perturbations in signal values at an unknown subset of nodes, when nothing is known about the nodes at which this corruption occurs. The corrupted nodes are also not assumed to be the same across different training observations. {\color{black} Our primary goal is to model the graph signal output $\mathbf{t}_n$ with a kernel regression model, so as to make predictions for the output when a new test input is given to the model. In order to achieve this, we start from the case of linear regression and proceed to derive the kernel regression case.
			\subsection{{\color{black}Linear basis model for regression} over graphs}
		\label{GLRsec}
		 The predicted output $\mathbf{y}_n$ of linear regression for $\mathbf{x}_n$ is given by
	\begin{equation}
	\mathbf{y}_n 
	=\mathbf{W}^\top\pmb\phi(\mathbf{x}_n),
	\label{eq:regression_output}
	\end{equation}
	where $\pmb\phi(\mathbf{x}_n)\in\mathbb{R}^K$ is some pre-defined function of $\mathbf{x}_n$ and $\mathbf{W}\in\mathbb{R}^{K\times M}$ denotes the regression coefficient matrix. {Equation \eqref{eq:regression_output} is referred to as \color{black} linear basis model for regression, usually termed linear regression for brevity (cf. Chapters 3 and 6 in \cite{Bishop}). Given the output model, we learn the optimal parameter matrix $\mathbf{W}$ by minimizing the following cost function:
	\begin{align}
	\label{eq:l1_cost}
	C_1(\mathbf{W})=\sum_{n=1}^N \|\mathbf{t}_n-\mathbf{y}_n\|_1 + \alpha \mbox{tr}(\mathbf{W}^\top\mathbf{W})+\beta \sum_{n=1}^N \mathbf{y}_n^\top\mathbf{L}\mathbf{y}_n,
	\end{align}
	where regularization coefficients $\alpha,\beta\geq 0$, $\mbox{tr}(\cdot)$ denotes the trace operator, and $\|\mathbf{x}\|_p$ the $\ell_p$ norm of $\mathbf{x}$. We minimize the $\ell_1$ norm of the model error since we assume that the corruptions occur due to presence of a sparse noise. The graph-Laplacian based regularization promotes the predicted output to be smooth over the given graph. 
	Finally, the regularization $\mbox{tr}(\mathbf{W}^\top\mathbf{W}) = \|\mathbf{W}\|_F^2$ ensures that $\mathbf{W}$ remains bounded. We refer to \eqref{eq:l1_cost} as the {\it linear regression over graphs for sparse noise} (LRGS). Although LRGS is a convex cost, it does not result in a closed-form solution. As a result, a direct kernel regression model is not possible for LRGS, since the kernel substitution or kernel trick cannot be employed in the absence of a closed form solution.
	%
	In this paper, we adopt the iteratively reweighted least squares approach (IRLS) to solve $\ell_1$ minimization problems described in the next Section. This is because IRLS helps  translate \eqref{eq:l1_cost} into a more general kernel regression approach.}
	\subsection{Iteratively reweighted least squares (IRLS)}
	\label{sec:irls}
	Iteratively reweighted least squares (IRLS) is a popular approach for solving $\ell_1$ minimization problems iteratively \cite{Daubechies_irls,irls_chartrand}. IRLS solves an $\ell_1$ minimization problem by successively solving a series of weighted least-squares problems, with weights that depend on values from the previous iteration. Consider the general problem for $\mathbf{z}^*=\arg\mathbf{z}\in\mathbf{R}^N$:
	\begin{equation*}
	\min\|\mathbf{z}\|_1, \,\,\mathbf{z}\in\mathcal{S}
	\end{equation*}
	where $\mathcal{S}\subset\mathbf{R}^N$ is the subspace in which $\mathbf{z}$ lies. Then, IRLS iteratively solves for optimal $\mathbf{z}$ as follows:
	\begin{eqnarray*}
	&&\mathbf{z}^*_n=\arg\min\|\mathbf{D}_n\mathbf{z}_n\|^2_2, \,\,\mathbf{z}\in\mathcal{S}\nonumber\\
	&&\mbox{s.t.  }\mathbf{D}_n=\mbox{diag}(d_{n,1},\cdots,d_{n,N} ),
	\end{eqnarray*}
	where $d_{n,i}=(z_{n-1}(i)+\delta)^{-1/2}$ and $\delta>0$ is a small constant introduced to circumvent ill-conditioning. In the case of convergence $\mathbf{z}_{n-1}\approx\mathbf{z}_{n}$ and then $\|\mathbf{D}_n\mathbf{z}_n\|^2_2=\sum_{i=1}^N z^2_n(i)z^{-1}_{n-1}(i)\approx\sum_{i=1}^N|z_n(i)|=\|\mathbf{z}\|_1$. Thus,  IRLS assymptotically solves an $\ell_1$ minimization problem.
	\subsection{Solving LRGS using IRLS}
	 We now propose to solve \eqref{eq:l1_cost} using IRLS by solving the following cost iteratively:
		\begin{equation}
	\label{eq:irls_cost}
	C(\mathbf{W})=\sum_{n=1}^N \|\mathbf{D}_n(\mathbf{t}_n-\mathbf{y}_n)\|^2_2 + \alpha \mbox{tr}(\mathbf{W}^\top\mathbf{W})+\beta \sum_{n=1}^N \mathbf{y}_n^\top\mathbf{L}\mathbf{y}_n,
	\end{equation}
	where $\mathbf{D}_n$ is the diagonal IRLS weighting matrix for the $n$th training observation such that
	\begin{equation} d_n(i)=|(t_{n-1}(i)-y_{n-1}(i))+\delta|^{-1/2}.
	\label{eq:d_irls}
	\end{equation}
	We define matrices $\mathbf{T}$, $\mathbf{Y}$ and $\mathbf{\Phi}$ as follows:
	\begin{eqnarray}
	\begin{array}{rcl}
	\mathbf{Y}&=&[\mathbf{y}_1\,\mathbf{y}_2\,\cdots \mathbf{y}_N]^{\top} \in\mathbb{R}^{N \times M},\\
	\mathbf{\Phi}&=&[\pmb\phi(\mathbf{x}_1)\,\,\pmb\phi(\mathbf{x}_2)\,\,\cdots \, \pmb\phi(\mathbf{x}_N)]^{\top} \in\mathbb{R}^{N \times K},\\
	\mathbf{T}_D&=&[\mathbf{t}_{1,D}\,\mathbf{t}_{2,D}\,\cdots \mathbf{t}_{N,D}]^{\top} \in\mathbb{R}^{N \times M},
	\end{array}
	\label{eq:matrix_definitions}
	\end{eqnarray}
	where $\mathbf{t}_{n,D}=\mathbf{D}_n^2\mathbf{t}_n$. 
	Using \eqref{eq:regression_output} and \eqref{eq:matrix_definitions}, the cost function \eqref{eq:irls_cost} is expressible as 
	\begin{eqnarray}
	\label{eq:cost_function_w}
	C_{}(\mathbf{W}) 
	& =  \displaystyle\sum_n \|\mathbf{D}_n\mathbf{t}_n\|_2^2-2 \, \mbox{tr}\left(\mathbf{T}_D^\top\pmb\Phi \mathbf{W}\right) \nonumber\\
	&\quad+ \mbox{tr}\left(  \sum_n \mathbf{W}^\top\pmb\phi(\mathbf{x}_n) \pmb\phi(\mathbf{x}_n)^\top\mathbf{W}\mathbf{D}_n^2\right) 
	\nonumber\\& \qquad\,\,\,\,+ \alpha\, \mbox{tr}(\mathbf{W}^\top\mathbf{W})+  \beta\, \mbox{tr}\left(  \mathbf{W}^\top\mathbf{\Phi}^\top\mathbf{\Phi}\mathbf{W}\mathbf{ L}\right).
	\end{eqnarray}
	 where we use properties of the matrix trace operation.
Since the cost function is quadratic in $\mathbf{W}$, we get the globally optimal and unique solution by setting the gradient of ${C}$ with respect to $\mathbf{W}$ equal to zero.
Setting $\displaystyle\frac{\partial C_{}}{\partial \mathbf{W}} =0$, we get that
	\begin{eqnarray}
&&-\mathbf{\Phi}^\top\mathbf{T}_D+\sum_n\pmb\phi(\mathbf{x}_n) \pmb\phi(\mathbf{x}_n)^\top\mathbf{W}\mathbf{D}_n^2+\alpha \mathbf{W} \nonumber\\&&\quad+\beta \mathbf{\Phi}^\top\mathbf{\Phi W L}= \mathbf{0}, 
	\mathrm{or,} \,\,\,\,\\
	&&\alpha\mathbf{W}+\beta \mathbf{\Phi}^\top\mathbf{\Phi W }\beta\mathbf{L}+\sum_n\pmb\phi(\mathbf{x}_n)\pmb\phi(\mathbf{x}_n)^\top\mathbf{W}\mathbf{D}_n^2=\mathbf{\Phi}^\top \mathbf{T}_D,\nonumber
	\label{eq:J_ww}
	\end{eqnarray}
	which on vectorizing and rearranging, gives
the optimal parameter matrix ${\mathbf{W}}_{\star}$ as:
	\begin{align}
	&\mbox{vec}({\mathbf{W}}_{\star}) \! \nonumber\\&= \!  \left[\!\alpha\mathbf{I}_{MN}\!  \! + \sum_n\mathbf{D}_n^2\otimes\pmb\phi(\mathbf{x}_n)\pmb\phi(\mathbf{x}_n)^\top\! +\beta\mathbf{L}\! \otimes \! \mathbf{\Phi}^\top\mathbf{\Phi})\!\right]^{-1}   \nonumber\\
	&\qquad\times\mbox{vec}(\mathbf{\Phi}^\top\mathbf{T}_D),\quad\nonumber
	\end{align}
	where $\mbox{vec}(\cdot)$ denotes the standard vectorization operator and $\otimes$ denotes the Kronecker product operation \cite{Loan1}.
	The predicted output for a new input $\mathbf{x}$ is then given by
	\begin{equation}
{\mathbf{y}}= {\mathbf{W}}_{\star}^{\top} \pmb\phi(\mathbf{x}).\nonumber
	\end{equation} 
Our development of linear regression  so far seemingly assumes that we have access to an input feature function $\pmb\phi(\cdot)$. In many practical cases, it is usually not clear what such a function should so that it helps achieve a reasonable prediction model through Eq \eqref{eq:regression_output}. On the other hand, it is generally more clear as to what kind of association is reasonable between various inputs. This can be usually captured in the form of a kernel function between input points, for example with radial-basis functions or Gaussian kernels. This makes kernel regression more intuitive and favourable in designing good predictors than linear regression. We next show that the approach we have pursued so far does not actually require an explicit $\pmb\phi(\cdot)$, since it only uses inner-products of the form $\pmb\phi(\mathbf{x}_m)^\top\pmb\phi(\mathbf{x}_n)$. Hence, the results may be reformulated completely in terms of kernel functions by substituting the inner product with general kernel functions $k(\mathbf{x}_m,\mathbf{x}_n)$-- thus leading to the kernel regression over graphs for sparse noise. 
	
	\subsection{Kernel regression over graphs for sparse noise (KRGS)}
We now demonstrate how IRLS approach to LRGS naturally leads to the more general kernel regression problem, that is, to kernel regression over graphs for sparse noise (KRGS).	
To achieve this, we use the substitution $\mathbf{W}=\mathbf{\Phi}^\top \mathbf{\Psi}$ and express the cost function in terms of the parameter $\mathbf{\Psi}\in\mathbb{R}^{N\times M }$. 
 This substitution is motivated by observing that on rearranging the terms in \eqref{eq:J_ww}, we get that
	\begin{equation}
	\mathbf{W} = \mathbf{\Phi}^\top\pmb\Psi,\nonumber
	\end{equation}
	where $ \pmb\Psi=\left[ \frac{1}{\alpha} \left( \mathbf{T}_D - \beta \mathbf{\Phi W L}   - \mathbf{\Phi}_D \right) \right]$, such that $\pmb\Phi_D\in\mathbb{R}^{N\times M}$ is the matrix whose $n$th row is given by $\pmb\phi(\mathbf{x}_n)^\top\mathbf{W}\mathbf{D}^2_n$.
	  On substituting $\mathbf{W}=\mathbf{\Phi}^\top \mathbf{\Psi}$ in Equation \eqref{eq:cost_function_w}, $\pmb\Psi$ becomes the parameter matrix for the dual cost:
	\begin{align}
		\label{J_dual}
	C(\mathbf{\Psi})=&-2\mbox{tr}\left(\mathbf{T}_D ^\top\mathbf{\Phi}\mathbf{\Phi}^\top\mathbf{\Psi}\right) \nonumber\\&+\mbox{tr}\left(  \sum_n \mathbf{\Psi}^\top\pmb\Phi\pmb\phi(\mathbf{x}_n) \pmb\phi(\mathbf{x}_n)^\top\pmb\Phi^\top\mathbf{\Psi}\mathbf{D}_n^2\right) \\
	& +\alpha \,\mbox{tr}(\mathbf{\Psi}^\top\bf{\Phi\Phi^\top}\mathbf{\Psi}) +  \beta\, \mbox{tr}\left( \mathbf{\Psi}^\top\bf{\Phi\Phi^\top}\bf{\Phi\Phi^\top}\mathbf{\Psi}\mathbf{ L}\right).\nonumber
	\end{align}
Kernel substitution or the kernel trick refers to the procedure in regression wherein all inner-products of feature vectors are substituted by a more general kernel function\cite{Bishop}. That is, one replaces all inner-products of the form $\pmb\phi(\mathbf{x}_m)^\top\pmb\phi(\mathbf{x}_n)$ with a kernel between inputs $\mathbf{x}_m$ and $\mathbf{x}_n$ given by $
k_{m,n}=k(\mathbf{x}_m,\mathbf{x}_n)$, where $k(\cdot)$ is some valid kernel function, such as the radial-basis function or the Gaussian kernel.
 Correspondingly, the matrix  $\mathbf{\Phi}\mathbf{\Phi}^\top\in\mathbb{R}^{N\times N}$ is generalized to the kernel matrix $\mathbf{K}$ between training samples such that its $(m,n)$th entry is given by $k_{m,n}$.
	Further let us denote $\mathbf{k}(\mathbf{x})=[k_1(\mathbf{x}), k_2(\mathbf{x}),\cdots, k_N(\mathbf{x})]^\top$ and $k_n(\mathbf{x})=k(\mathbf{x}_m,\mathbf{x})$. Then, we have that
		\begin{align}
	\label{J_dual}
	C_{}(\mathbf{\Psi})
	=&-2\mbox{tr}\left(\mathbf{T}_D^\top\bf{K}\mathbf{\Psi} \right) +\mbox{tr}\left(  \sum_n \mathbf{\Psi}^\top\mathbf{k}(\mathbf{x}_n) \mathbf{k}(\mathbf{x}_n)^\top\mathbf{\Psi}\mathbf{D}_n^2\right)\nonumber\\
	& +\alpha\, \mbox{tr}(\mathbf{\Psi}^\top\bf{K}\mathbf{\Psi}) +  \beta \,\mbox{tr}\left( \mathbf{\Psi}^\top\bf{K}\bf{K}\mathbf{\Psi}\mathbf{ L}\right),
	\end{align}
	{\color{black} In kernel regression literature, \eqref{J_dual} is termed a dual representation of Equation \eqref{eq:cost_function_w} (cf. Chapter 6 of \cite{Bishop}).} Thus, we see that the entire regression problem is generalized to the usage of  arbitrary kernel functions, without the need of an explicit input feature $\pmb\phi$.
Setting the derivative of $C(\mathbf{\Psi})$ with respect to $\mathbf{\Psi}$ to zero, we get that
		\begin{align}
	\label{J_dual_deriv}
	\mathbf{0}=\frac{\partial C_{}(\mathbf{\Psi})}{\partial\mathbf{\Psi}}
	&=-2\mathbf{K}\mathbf{T}_D\mathbf{\Psi} +2 \sum_n \mathbf{k}(\mathbf{x}_n) \mathbf{k}(\mathbf{x}_n)^\top\mathbf{\Psi}\mathbf{D}_n^2 \nonumber\\ &\quad+2\alpha\,\mathbf{K}\mathbf{\Psi} +2\beta \mathbf{K}\mathbf{K}\mathbf{\Psi}\mathbf{ L}.
	\end{align}
	On vectorizing both sides of \eqref{J_dual_deriv} and rearranging , we get that
	\begin{align}
&\mbox{vec}(\mathbf{\Psi})=
	\mathbf{B}\mbox{vec}(\mathbf{T}_D),\,\,\mbox{where}\nonumber
\\
	\mathbf{B} &=  \left[\sum_n \mathbf{D}_n^2\otimes \mathbf{k}(\mathbf{x}_n) \mathbf{k}(\mathbf{x}_n)^\top+\alpha(\mathbf{I}_M\otimes \mathbf{K})+(\beta \mathbf{L}\otimes \mathbf{K}^2)\right]^{-1}\nonumber\\
	&\quad\times(\mathbf{I}_M\otimes \mathbf{K}).
	\label{eq:B}
	\end{align}	
	%
	%
	Using the $\pmb\Psi$ so computed, the output of the kernel regression for a new test input $\mathbf{x}$ is given by
	\begin{align}
	\begin{array}{rcl}
	\mathbf{y}&=&\mathbf{W}^{\top} \, \pmb\phi(\mathbf{x}) =\mathbf{\Psi}^{\top} \, \mathbf{\Phi}\pmb\phi(\mathbf{x})	= \mathbf{\Psi}^{\top}\mathbf{k}(\mathbf{x}) \\
	& = & \left(\mbox{mat} \left( \mathbf{B}\mbox{vec}(\mathbf{T}_D) \right) \right)^{\top}\mathbf{k}(\mathbf{x}),
	\end{array}
	\label{eq:KRG_output}
	\end{align}
	where $\mathbf{k}(\mathbf{x})=[k_1(\mathbf{x}), k_2(\mathbf{x}),\cdots, k_N(\mathbf{x})]^\top$. Here $\mbox{mat}(\cdot)$ denotes the reshaping operation of an argument vector into an appropriate matrix of size $N \times M$ by concatenating subsequent $N$ length sections as columns. The complete process of KRGS is summarized in Algorithm 1.
	\begin{algorithm}[t]
	\caption{ Kernel Regression over Graphs for Sparse Noise}
	\begin{algorithmic}[1]
			\State Initialize $\mathbf{D}_n=\mathbf{I}_M$ for all $1\leq n \leq N$, set $i_{max}$ maximum number of iterations 
			\While {$i< i_{max}$} 
			\State Compute $\pmb\Psi=\mathbf{B}\mbox{vec}(\mathbf{T}_D)$ where
			 $\mathbf{B}$ is as given in \eqref{eq:B}.
			\State For input $\mathbf{x}$, output predicted $\mathbf{y}=\mathbf{\Psi}^{\top} \mathbf{k}(\mathbf{x})$.
			\State $\mathbf{D}_n=\mbox{diag}(d_n(1),d_n(2),\cdots,d_n(M))$ where $d_n(m)$ is given by \eqref{eq:d_irls}.
				\State $i\to i+1$
			\EndWhile
	\end{algorithmic}
\end{algorithm}
	
	\section{Experiments}
	\begin{figure}[t]
		\centering
		$
		\begin{array}{cc}
		\hspace{-.1in}
		\subfigure[]{	\includegraphics[width=1.5in]{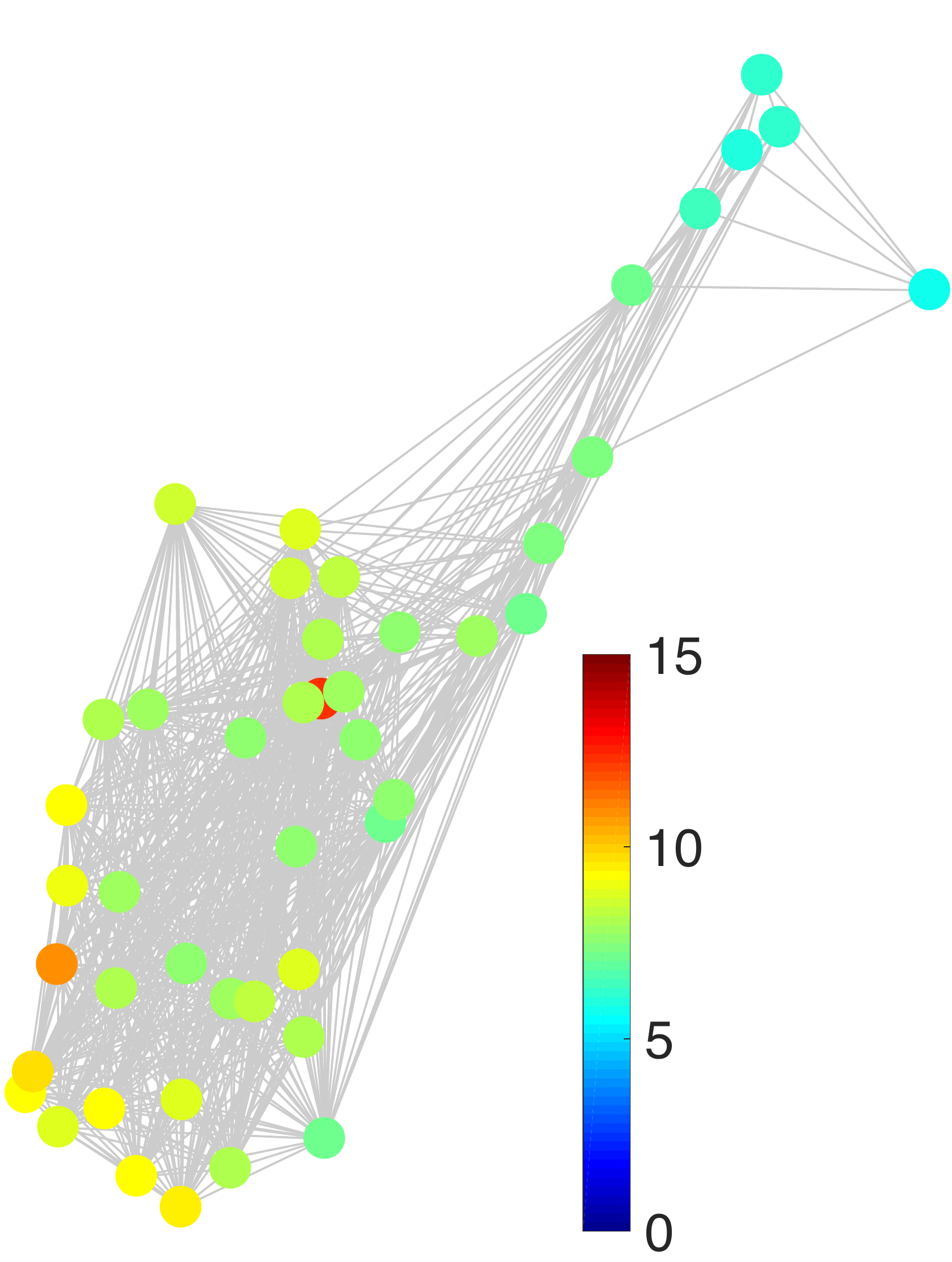}}
		\hspace{-.0in}\\
		\subfigure[]{	\includegraphics[width=1.5in]{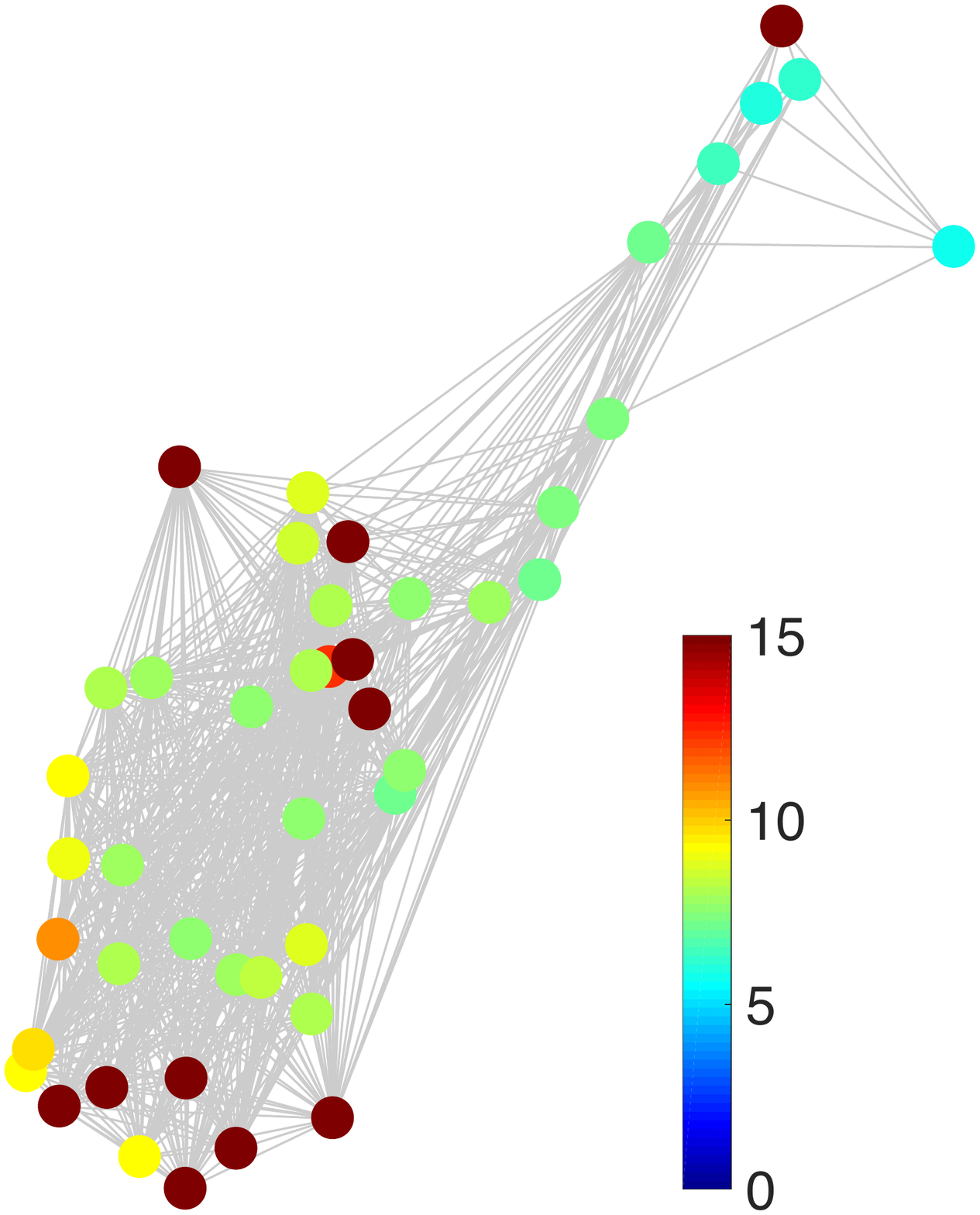}}
		\hspace{-.08in}
		\subfigure[]{	\includegraphics[width=1.5in]{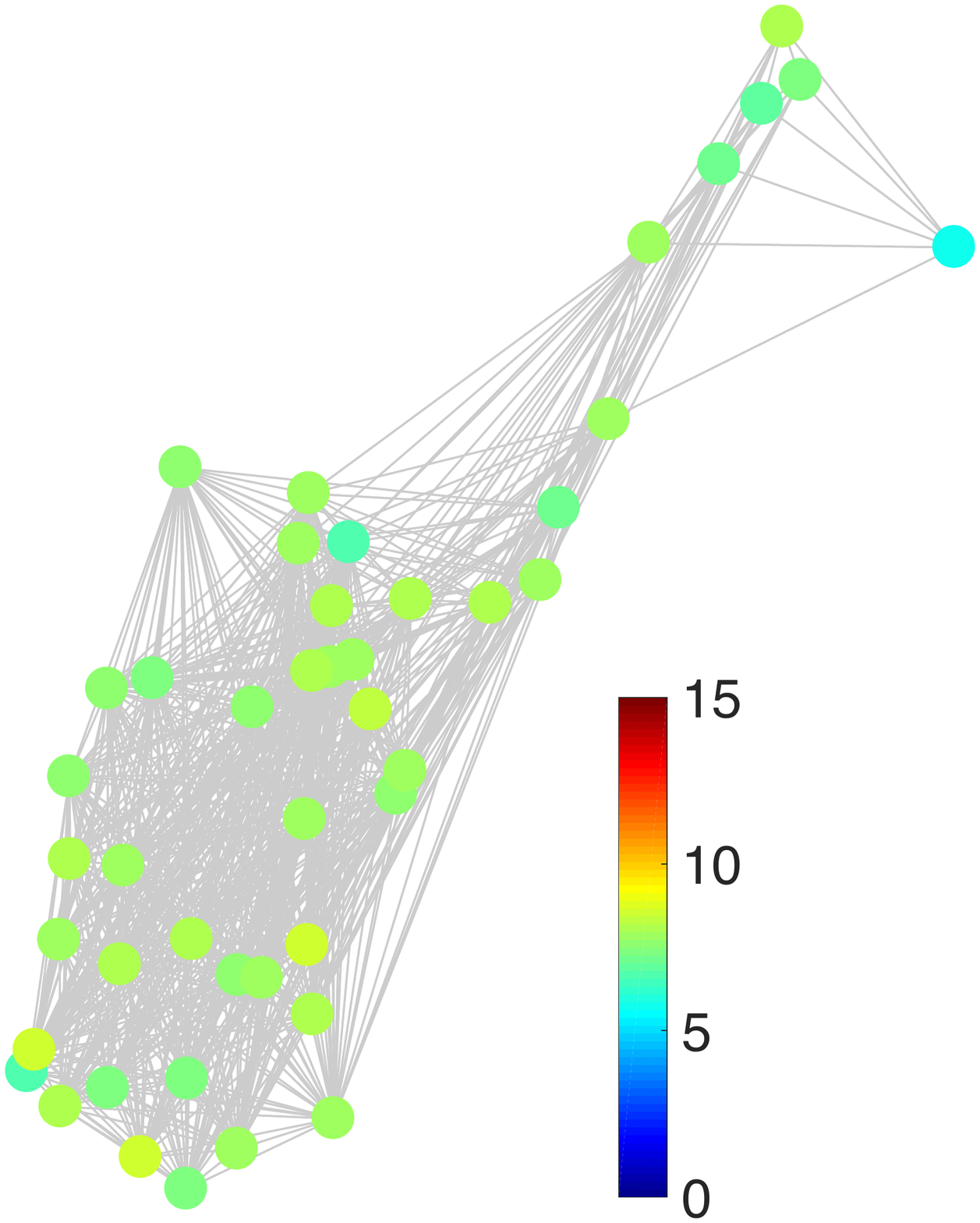}}
		\end{array}$
		\caption{Example showing result of our approach on temperature data over Sweden on training data: (a) True output (b) Output with sparse noise, and (c) Predicted output from our approach. The colorbar shows the temperature scale in degree celsius. 
		}
		\label{l1_krg:introfig}
	\end{figure}
	We consider the application of our approach to real-world graph signal of temperature measurements taken over 45 most populated cities in Sweden for the period of three months from September to November 2017 \footnote{The data is available publicly from the Swedish Meteorological and Hydrological Institute \cite{SMHI}.}. Our hypothesis is that our approach achieves good prediction performance in the two cases of corruptions in the graph signal training data: missing samples, and large perturbations.
	
	 The input $\mathbf{x}\in\mathbb{R}^{45}$ is the vector of temperature measurements of all 45 cities of one day, and the output  $\mathbf{t}_n\in\mathbb{R}^{45}$ to be predicted contains the temperature measurements of the 45 cities for the succeeding day. This gives us a total of 92 input-output pairs. We use the first $N_{tr}=46$ pairs for training, and the remaining $N_t=46$ for testing. We consider the graph adjacency matrix constructed using geodesic distances between the cities as follows:
$\mathbf{A}(i,j)=\displaystyle\exp{\left(-\frac{d_{ij}^2}{\sum_{i,j}d_{ij}^2}\right)},\nonumber
$
where $d_{i,j}$ denotes the geodesic distance between the $i$th and $j$th cities. 
We measure the prediction performance on test data in terms of the normalized mean square error (NMSE):
\begin{equation}
\mbox{NMSE}=10\log_{10}\left(\frac{\mathrm{E}\left(\sum_{n=1}^{Nt}\|\mathbf{y}_n-\mathbf{t}_{0,n}\|_2^2\right)}{\mathrm{E}(\sum_{n=1}^{Nt}\|\mathbf{t}_{0,n}\|_2^2)}\right),\nonumber
\end{equation}
where $\mathbf{y}_n$ denotes the regression output and $\mathbf{t}_{0,n}$ the true value of output which does not contain any noise, and $Nt$ denotes the number of test data points. The experiment is performed for different training data sizes of $N$ drawn from the full training set of $N_{tr}=46$ samples. The parameter $\delta$ is set experimentally to $0.1$. The parameters $\alpha$ and $\beta$ are computed through four-fold cross validation. We use Gaussian kernel in all our experiments whose parameter $\sigma$ is also set using four-fold cross-validation.
We initialize our algorithm with $\mathbf{D}_n$ as the identity matrix for all $N$ samples. This corresponds to solving the cost with $\ell_2$ norm of the model error as the first iteration. The regression problem with such a cost was defined in \cite{Arun_kergraph}, where it was termed as the kernel regression over graphs (KRG). We compare the performance to the baseline performance obtained using KRG as the iterations increase. This gives us a clear picture of the gain in using KRGS over KRG. We note here that KRG has been shown to outperform kernel ridge regression (KRR)\cite{kergraph1}, a state-of-the-art approach in graph signal reconstruction/inpainting which employs kernels across nodes \cite{Arun_kergraph}. Since KRR was reported to outperform other competing approaches, it suffices to make comparisons only with KRG.

We simulate the missing samples as follows: For each training output graph signal, we randomly choose $25\%$ of the nodes and set their signal values to equal to zero. The subset of nodes where the signal values are missing is independent from one training observation to the other. In the case of large perturbations, we adopt the same strategy but instead of setting signal values to zero, we scale them by a constant factor of 4. This experimentally results in a training samples to have a signal-to-noise-ratio (SNR) of in the range of 6 to 10dB in both the sparse noise cases. An example of application of our approach with large perturbations is illustrated in Figure \ref{l1_krg:introfig}. In Figure \ref{l1_krg:missing}, we show the NMSE as a function of the iterations for various training sample sizes $N$, averaged over 100 Monte Carlo simulations of missing samples or perturbations. We observe that for each training sample size $N$, the NMSE reduces with the number of iterations, and we find that convergence is acheived after five to ten iterations. We also observe that our approach yields significant gain in prediction performance compared to KRG, which corresponds to the solution obtained at the first iteration of our approach. We also note that while we have made use of a single $\delta$ in all our experiments, changing $\delta$ with $N$ could possibly help achieve even better prediction performance.

\begin{figure}
	\centering
	$
	\begin{array}{cc}
	\hspace{-.125in}
	\subfigure[]{	\includegraphics[width=1.7in]{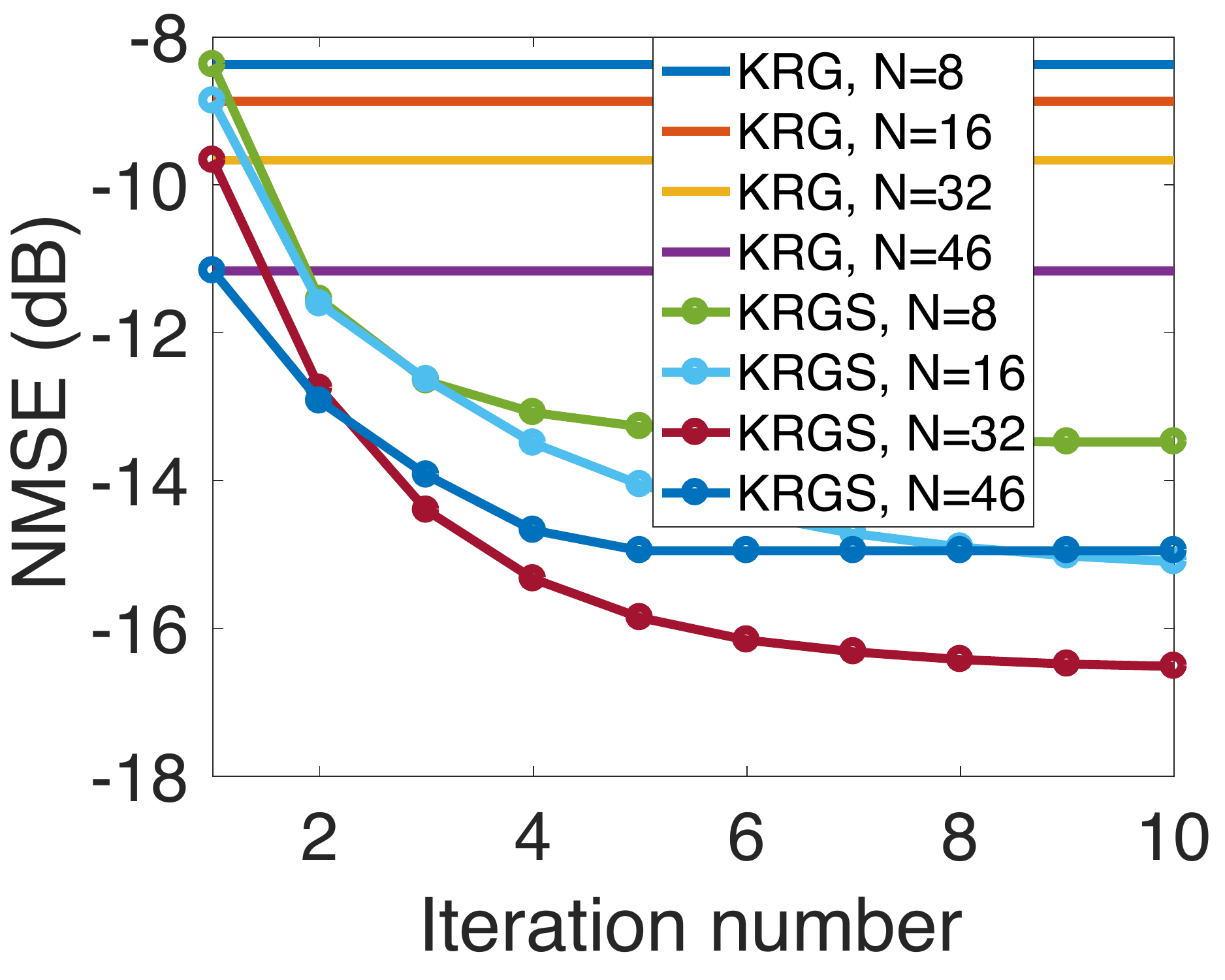}}
	\hspace{-.0in}
	\subfigure[]{	\includegraphics[width=1.7in]{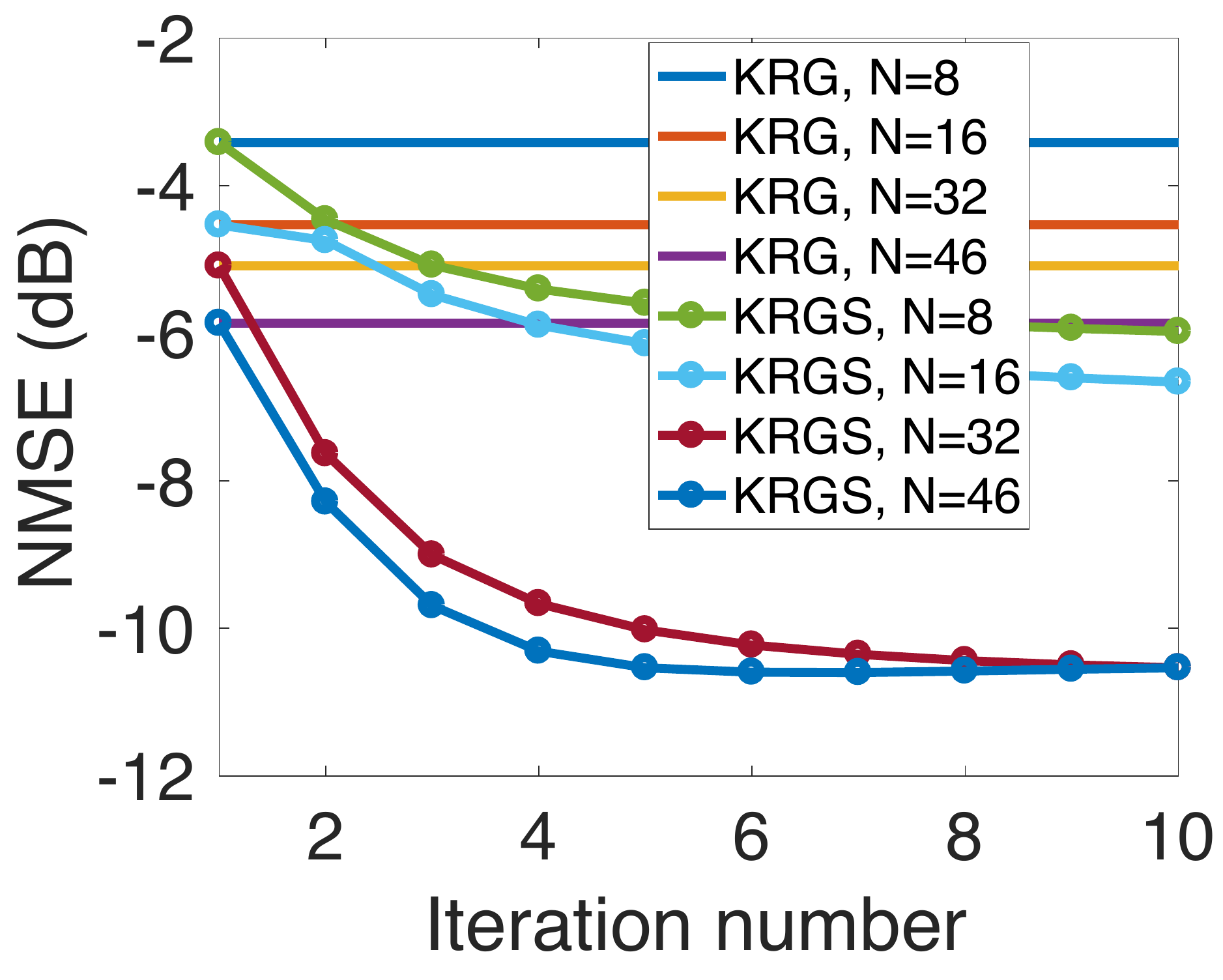}}
	\end{array}$
	\caption{Results for temperature prediction experiment. NMSE as a function of number of iterations, at different training sample sizes for (a) missing data, and (b) Large perturbations.}
	\label{l1_krg:missing}
\end{figure}

\section{Conclusions}
We proposed linear regression for predicting graph signals for the adverse scenario when the training data is corrupted with sparse noise, by posing it as a regularized $\ell_1$-norm minimization problem. We employed the iteratively reweighted least-squares approach to solve the optimization problem, which resulted in a closed form solution. This in turn enabled the development of the more general kernel regression for sparse noise scenario, through the use of kernel substitution. Application on a real-world graph signal dataset showed that our approach outperforms the $\ell_2$-norm based kernel regression by a very significant margin, and that the performance improves with the number of iterations. 
\newpage
	\bibliographystyle{IEEEtran}
	\bibliography{refs,refs_krg,refs_multikernel,refs_l1_krg}

\end{document}